\documentclass[journal]{IEEEtran}

\usepackage{graphicx}
\usepackage{multirow}
\usepackage{longtable}
\usepackage{rotating}
\usepackage{color}
\usepackage{makecell}
\usepackage{hyperref}
\usepackage{cite}
\usepackage{lineno}

\graphicspath{{figures/}}

\ifCLASSINFOpdf
\else
\fi
\begin{document}
%
\title{A Symmetric Encoder-Decoder with Residual Block for Infrared and Visible Image Fusion}
%
%
%

\author{Lihua~Jian,~
        Xiaomin~Yang,~
        Zheng~Liu,~
        ~Gwanggil~Jeon,
        ~Mingliang Gao,
        ~and~David Chisholm

\thanks{L.~Jian is with College of Electronics and Information Engineering, Sichuan University, Chengdu, Sichuan, 610064 P.R.China. He is currently as a visiting Ph.D. student with School of Engineering, University of British Columbia, V1V 1V7, Kelowna, BC, Canada (e-mail: jianlihua123@126.com).}
\thanks{X.~Yang (*Corresponding author*) is with College of Electronics and Information Engineering, Sichuan University, Chengdu, Sichuan, 610064 P.R.China (e-mail: arielyang@scu.edu.cn).}
\thanks{Z.~Liu and D.~Chisholm  are with School of Engineering, University of British Columbia, V1V 1V7, Kelowna, BC, Canada (e-mail: zheng.liu@ubc.ca, david.chisholm@alumni.ubc.ca).}

\thanks{G.~Jeon is with School of Electronic Engineering, Xidian University, Xi'an, 710071, China. And Department of Embedded Systems Engineering, Incheon National University, Incheon, 22012, Korea. (e-mail: ggjeon@gmail.com).}

\thanks{M.~Gao is with School of Electrical and Electronic Engineering, Shandong University of Technology, Zibo 255000, China (e-mail: mlgao@sdut.edu.cn).}

\thanks{This regular article is supported by China Scholarship Council (Grant No. 201806240047).}}


%
%

\markboth{~~~}%
{Jian \MakeLowercase{\textit{et al.}}: A Symmetric Encoder-Decoder with Residual Block for Infrared and Visible Image Fusion}
%



\maketitle

\begin{abstract}
In computer vision and image processing tasks, image fusion has evolved into an attractive research field. However, recent existing image fusion methods are mostly built on pixel-level operations, which may produce unacceptable artifacts and are time-consuming. In this paper, a symmetric encoder-decoder with residual block (SEDR) for infrared and visible image fusion is proposed. For the training stage, the SEDR network is trained with a new dataset to obtain a fixed feature extractor. For the fusion stage, first, the trained model is utilized to extract the intermediate features and compensation features of two source images. Then, extracted intermediate features are used to generate two attention maps, which are multiplied to the input features for refinement. In addition, the compensation features generated by the first two convolutional layers are merged and passed to the corresponding deconvolutional layers. At last, the refined features are fused for decoding to reconstruct the final fused image. Experimental results demonstrate that the proposed fusion method (named as SEDRFuse) outperforms the state-of-the-art fusion methods in terms of both subjective and objective evaluations.
\end{abstract}


%
\IEEEpeerreviewmaketitle

\section{Introduction}
%
%
%
%
\IEEEPARstart{I}{mage} fusion, as a promising technique for computer vision field, can be leveraged in many real applications. Generally, it can be devoted to object detection in night environments, disease diagnoses in the medical system \cite{bhatnagar2013directive}, photography \cite{guo2019fusegan,wang2015pseudo,kou2018intelligent} and remote-sensing for mapping, etc. Image fusion is designed to obtain a more comprehensive and informative image by integrating multiple source images from various sensors.  Recently, multi-sensor information such as thermal infrared and visible images has been widely applied to the surveillance areas on both military and civilian use \cite{hu2017adaptive,zhao2018multisensor}. In general, the infrared images are derived from thermal radiation of the objects, whereas visible images are merely captured from the visual scene. However, each of them has its limitations at the night vision as a single sensor cannot capture the complete information from a scene. Therefore, it is essential to fuse the multi-sensor data to generate an informative image, which can provide the final users with more complementary information.

\begin{figure}[h]
	\centering
	\includegraphics[width=9cm]{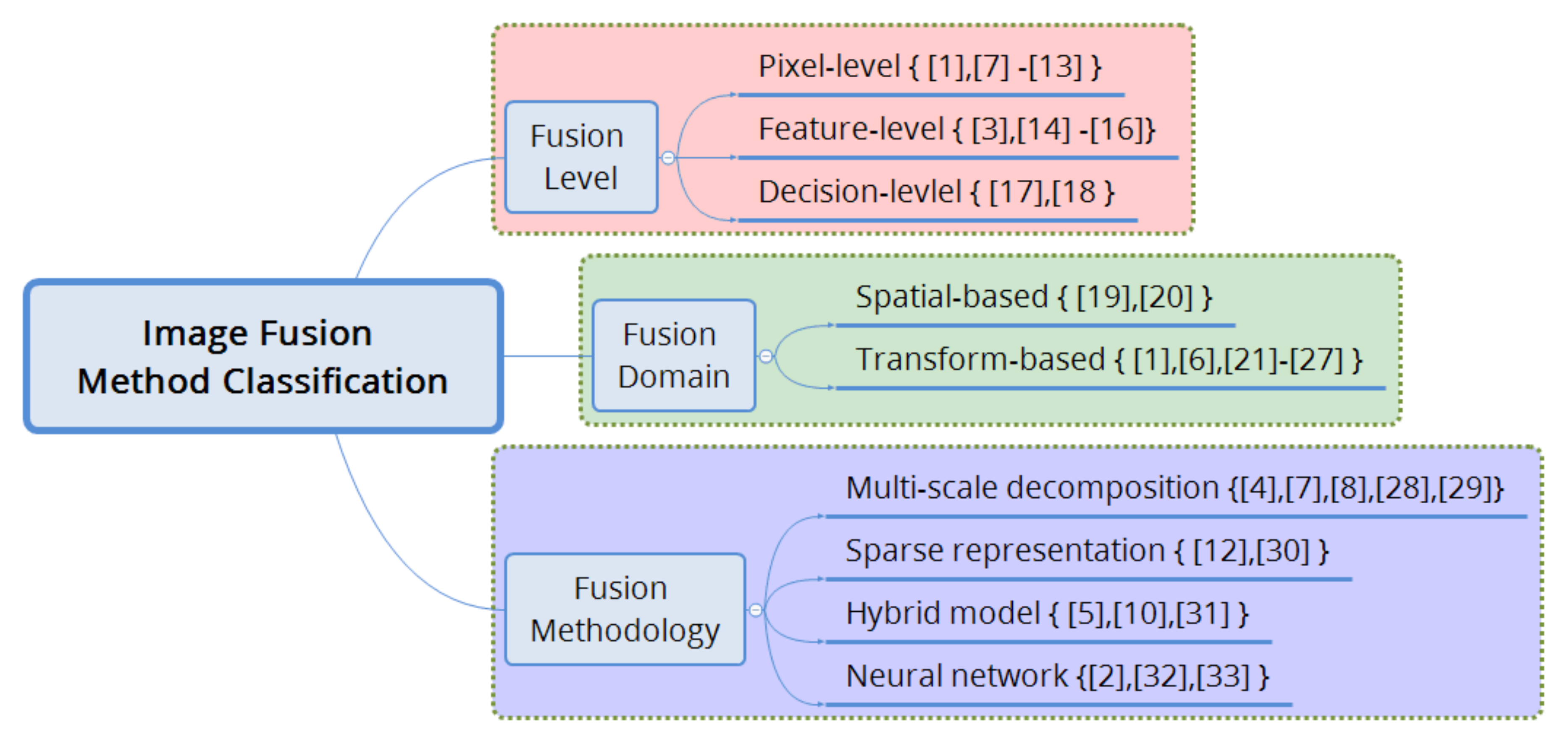}
	\caption{Classification for image fusion methods from various perspectives}
	\label{imgclass}
\end{figure}

Numerous image fusion algorithms for infrared and visible images have been proposed in these years. We can review these fusion methods from three perspectives: fusion level, fusion domain, and fusion methodology, shown in Fig.~\ref{imgclass}. From the implementation level, current fusion algorithms can be performed at three main levels, i.e., pixel-level, feature-level and decision-level. As the lowest level fusion techniques \cite{zhu2018novel,cao2015multi,liu2015general,ma2016infrared,jian2018multi,nejati2015multi,bhatnagar2015new}, the pixel-level based image fusion directly deals with the pixels of an image obtained from sensors. It aims to retain more original information of the source images for visual performance. However, in real applications, pixel-level fusion has two limitations. First, it needs to take more time to preprocess amounts of information. Second, it may result in serious degradation or distortion in the fused results without strict registration of the source images. Therefore, feature-level image fusion \cite{shao2018remote,li2018densefuse,liu2015multi} has become a promising direction with the development of deep learning techniques. This operation commonly extracts the most representative features of the source images by using specific filters or some other representation learning methods. Finally, the fused image is reconstructed via combining the useful features. Although this kind of fusion method misses certain information during the process, it has advantages in real-time processing applications such as object detection tasks. Decision-level fusion, the highest level fusion mechanism \cite{kolekar2016decision,kausar2016random}, is mainly based on machine learning. There are many challenges due to the modality correlation between different source images. It provides terminal users decisions instead of visual perception. Namely, it does not identity specific objects in specific images. Hence, it may not be suitable for most current computer vision tasks.

In addition, according to the fusion domain, existing image fusion methods can be roughly classified into two main aspects, typically, the spatial- and transform- based methods. Spatial-based fusion methods directly implement on the pixels of the source images. Examples of spatial-based fusion methods are weighted average, morphological  operations \cite{li2015multi}, and other matrix-computation methods \cite{wei2015fast}. Fused results by these methods usually produce undesired effects such as spectral distortions due to unbalanced transfer between the source images. In contrast, transform-based image fusion methods can avoid some limitations such as spectral degradation. By using the appropriate transformation tools, these methods are designed to project source images into various transformed components. The common transformation tools involve pyramid \cite{du2016union}, wavelet \cite{pajares2004wavelet,sappa2016wavelet}, curvelet \cite{dong2015high}, shearlet \cite{singh2015nonsubsampled,yin2017novel}, and contourtlet \cite{ganasala2014ct}, etc. Subsequently, different kinds of suitable fusion strategies are applied to merge the new transformed components. Finally, these fused components are reconstructed by an inverse transform to the original space domain.

In terms of the methodology, recent image fusion methods have been grouped into four major families: (1) the multi-scale decomposition or transformation (MSD or MST); (2) the sparse coding or representation (SR); (3) the hybrid fusion model (HFM); (4) the neural network (NN)-based.

Most of the existing fusion methods in recent years lean on the multi-scale decomposition framework \cite{jian2018multi,cui2015detail,zhu2018novel,bavirisetti2016two}. Typical operations of the MSD are as follows: (1) various represented layers are extracted at different scales (including dual-scale) by the specific transform tools such as pyramid, wavelet, and edge-preserving filters;  (2) these extracted feature layers or transform coefficients are combined together by specific fusion rules; (3) previously obtained fusion layers are summed up or inverse transformed to obtain the final fused image. For example, literature \cite{zhu2018novel} decomposed the source images into cartoon and texture components, which are fused by using Sum-modified-Laplacian (SML) and sparse representation fusion strategies respectively.  

Recently, the sparse coding method is applied to various signal processing fields \cite{yang2008image,nejati2015multi} to deal with two-dimensional images.  The sparse representation learning may be considered as the best feature-representation approach.  Researchers encode the source images on an over-complete dictionary to obtain the sparse coefficients, which can be fused by using different fusion strategies such as the $l_{1}$-norm, choose-max, and weighted average, etc, \cite{nejati2015multi}. The final fused image is restored by using these merged coefficients on the same over-complete dictionary. In addition, the SR-based method can also serve as a fusion strategy. 

To fully combine the special merits of each fusion method, a few hybrid fusion models (HFMs) \cite{liu2015general,cai2017fusion} have been successfully explored. An example of hybrid models combines the multi-scale transform (MST) and sparse representation (SR) \cite{liu2015general}. This hybrid model aims to extract useful information such as low-frequency sub-band features via the MST tool. However, using weighted average and choose-max strategies for low-frequency integration will result in redundant information (visual artifacts) on the fused image, since the low-frequency components of an image indicate energy. At present, it has been proven that the use of SR techniques can express the energy without considering over-smoothing. To some extent, it can reduce the redundant information and improve the visual performance of the fused image. 

Currently,  the most promising and attractive direction for image fusion is the deep learning-based methods. It is not enough in the field of image fusion even though some novel successful cases \cite{ma2019fusiongan,liu2017multi} have been presented. Neural networks (NNs), such as the autoencoders (AEs) with their variants, deep belief networks (DBNs), and convolutional neural networks (CNNs), are the core of deep learning techniques. All of these NN-based methods are built on training a group of weights that are similar to a series of filters. Then, these trained weights are applied to extract different types of feature layers. Therefore, how to select a training model and how to use their abstract information effectively is critically important to fusion results. Liu et al. \cite{liu2017multi} proposed a CNN-based fusion method for multi-focus images. They used CNNs to identify clear and unclear parts of multi-focus images. In fact, this is a binary classification problem. However, this framework has no generalization capabilities in other image fusion applications such as infrared and visible images. Ma et al. \cite{ma2019fusiongan} presented a FusionGAN-based fusion method for infrared and visible images. This paper proposed a novel method to fuse two types of information using a generative adversarial network (GAN). However, this network has changed the original information of the source images to some extent.

Most existing image fusion methods are based on the pixel level, which may encounter two key issues, namely, it is time-consuming and generates redundant information. 

In this paper, to overcome the time-consuming and quality degradation of the fused image, a novel deep learning method for infrared and visible image fusion is proposed. First, a symmetric encoder-decoder with residual block (SEDR) network is trained with an available datasets (KAIST and FLIR) including infrared and visible images. Then, the trained model is utilized to extract the intermediate features and compensation features of the source images. Subsequently, the intermediate features are exploited to generate the corresponding weight maps by using a softmax function. Multiplying the features with the corresponding weight maps to obtain two attention maps, are then employed to fuse the intermediate features. Additionally, the obtained compensation features need to be merged first and passed to the corresponding deconvolutional layers. Finally, all the merged features are fed back into the decoder part by deconvolution operations to reconstruct the final fused image.  

The main contributions of the proposed fusion framework can be enumerated as follows:
\begin{itemize}
	\item propose a symmetric encoder-decoder with residual block network. 
	\item present an attention map-based feature fusion method. 
	\item apply the skip connections to compensate for the missing details of reconstruction images. 
	\item train the proposed network on KAIST and FLIR datasets consisting of infrared and visible images. 
\end{itemize}

The rest of this article is arranged as follows. Section 2 reviews related work in our fusion framework. Section 3 details the fusion framework. Section 4 interprets the experiments and discussion. Section 5 concludes this article.

\section{Related Work}

\subsection{CNNs-based Work Review}
In recent years, deep learning techniques have shown great advantages in many image processing applications, i.e., image super-resolution, image segmentation, and object detection. Liu et al. \cite{liu2017multi} applied convolutional neural networks (CNNs) to multi-focus image fusion for the first time. They demonstrated the feasibility of CNN for image fusion. However, it simulates the negative examples (defocus images) by Gaussian blur, which makes the training dataset impractical \cite{ma2019fusiongan}. In addition, this model cannot be generalized to other multi-modality image fusions. For example, like infrared and visible images, it may be inappropriate to treat them as a classification problem. Moreover, there are no sufficient ground-truth images for supervised training networks.

Unlike supervised training, unsupervised learning methods commonly use themselves as targets (labels) to train a network, which compensates for insufficient labeled training data. For instance, an autoencoder architecture can compress the input data into a latent representation (also known as a hidden feature) and then restore the original input data with a low reconstruction error. The ground-truth of this restoration is the input data. Haeggstroem et al. \cite{haeggstroem2018deeprec} leveraged a deep encoder-decoder network to solve the inverse problem of PET reconstruction, which quickly and directly obtained high-quality images with little noise information from the PET sinogram data. A stacked convolutional denoising auto-encoder for feature representation was presented by Du et al. \cite{du2017stacked}, which could learn a powerful feature extractor. Note that, an encoder-decoder architecture has good reconstruction characteristics without supervised learning. More research attention should focus on the content and use of features, especially the feature-level fusion of multi-modality images.   

Inspired by previous work, we intend to use the restoration ability of the encoder-decoder network to obtain a fixed feature extractor, learning a hierarchy of complex representation features. Considering that these features convey different types of information from source images, the framework is divided into three parts at the fusion stage. To begin with, we train a network to have a low reconstruction loss on the specific dataset. Then, the encoder part of the trained network can serve as a feature extractor. In contrast, the decoder part can be regarded as a generator for image reconstruction. The features obtained by each convolutional layer can be fused by using related fusion strategies, and the fused features are transferred to the corresponding deconvolutional layers of the decoder part. To our best of knowledge, this is the first time that all  features generated by convolutional layers are considered for fusion, which fully retains each level of information in the fused image.        


\subsection{Symmetric Encoder-Decoder with Residual Block}

Many image restoration algorithms have been built on neural networks (NN), especially based on deep convolutional encoder-decoder networks. These encoder-decoder networks have superior performance due to their strong ability to learn the intermediate hidden information, which can be used to restore the original image with denoising.

\begin{figure*}[h]
	\centering
	\includegraphics[width=17cm]{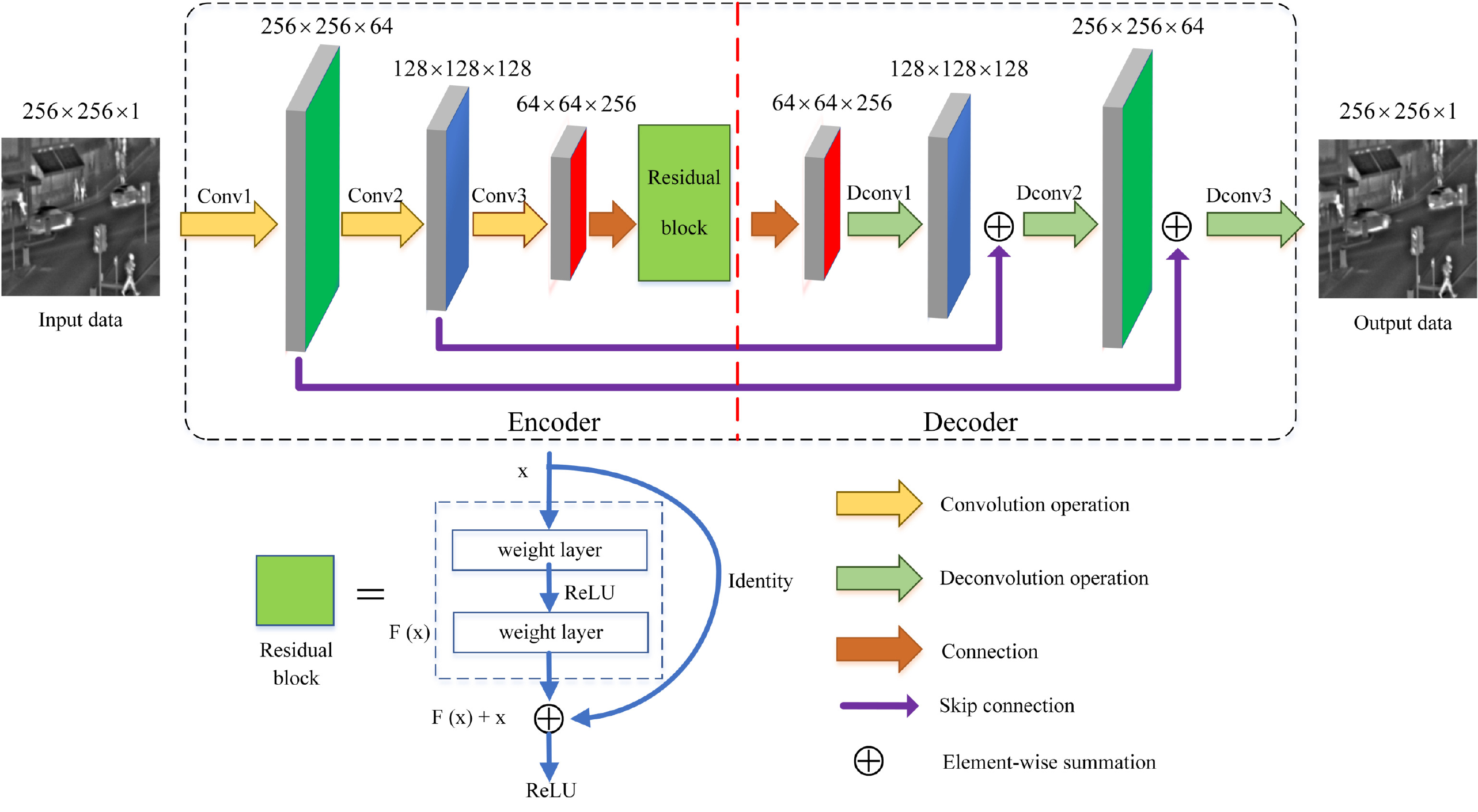}
	\caption{Overall training architecture: a symmetric encoder-decoder with residual block (SEDR) network. }
	\label{train}
\end{figure*}

In this work, a symmetric encoder-decoder with residual block (SEDR) network is trained with the specific dataset including infrared and visible images, as shown in Fig.~\ref{train}. The goal of this training framework is to accurately reconstruct the original dataset (raw input data) while minimizing the reconstruction loss. That is, the smaller the reconstruction error, the more representative the extracted features are. Our proposed SEDR network involves two parts at the training stage. The two major parts are an encoder and decoder without a fusion layer. Once the feature extractor has been trained, we will add the fusion part to complete the framework (see Fig.~\ref{frame}). The basic units in the SEDR network are convolutional layer, deconvolutional layer, residual block \cite{he2016deep}, skip connections and rectified linear units (ReLU) function. The pooling layer is removed because some useful details from the original dataset will be lost. 

\textbf{Encoder part.}~~Our encoder part consists of three convolutional layers and one residual block.  The first convolutional layer does not change the input size, while the second and the third convolutional layers (down-sampling) are half the size. All convolution operations act as feature extractors, fully retaining texture and structural information of the source images. To compensate for the missing image details during the convolution process, we mimic the ResNet \cite{he2016deep} to further reuse the previous features. In this network, we add one residual block after the last convolutional layer. The input training data and the output generation is of same size $256 \times 256 \times 1$ (height, width, channel), and the output from the encoder has 256 intermediate feature maps with the size of $64 \times 64$, retaining more raw structural details.

\textbf{Decoder part.}~~To obtain the output image of the same size as the input, the decoder part adopts a symmetric deconvolution to corresponded convolutions in the encoder part. Deconvolution is usually used to reconstruct the original image from extracted intermediate features by up-sampling. The kernel size of the deconvolutional layers must be the same with the convolutional layers to match exactly. In this network, all kernel size is set to $3 \times 3$.  Besides, the decoder part only has two types of units, deconvolutional layers and ReLU functions.

\textbf{Skip connections.} As depicted in the literature \cite{mao2016image}, the convolutional operations preserve the primary image content while the texture details of the image may be lost. In addition, deconvolution only can restore the structural details of image contents from the extracted features, which have a certain amount of information loss during down-sampling in the encoder part. In general, the output of the decoder is the filtered version of the input image, which results in unsatisfactory performance for image fusion. Therefore, in our work, we use skip connections to transfer texture feature information from convolutional layers to their corresponding deconvolutional layers by an element-wise, choose-max manner. These skip connections make the proposed framework easier to be trained and speed up the convergence. More training details will be discussed in Section~\ref{T}. 

\section{The Proposed Fusion Framework}

\begin{figure*}[h]
	\centering
	\includegraphics[width=17cm]{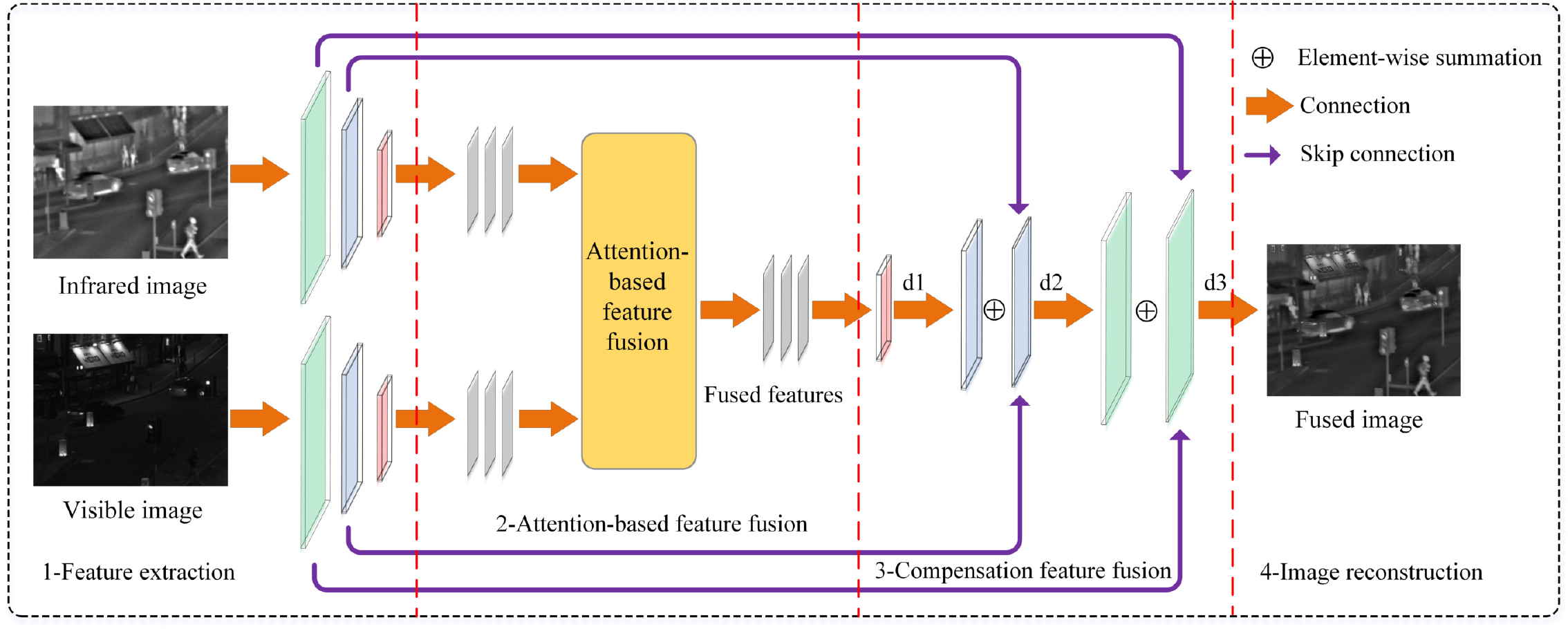}
	\caption{The proposed fusion framework (SEDRFuse).}
	\label{frame}
\end{figure*}

Fig.~\ref{frame} shows the proposed fusion framework, which is composed of four main phases: (1) feature extraction; (2) attention-based feature fusion; (3) compensation feature fusion; (4) image reconstruction. First, the source images are separately encoded by utilizing the trained model to extract a series of rough features (intermediate features). These intermediate features are further calculated to generate two different attention maps, which are used to refine these intermediate features in turn. Subsequently, an attention-based fusion strategy is applied to combine these refined features. Meanwhile, the previous features (compensation features) generated by the first and second convolutional layers are merged by using an element-wise summation, choose-max strategy. Finally, all the final fused features are fed back into the decoder part to reconstruct the final fused image.   

\subsection{Feature Extraction}
Image features should be learned from a specific image dataset that is similar to the characteristics of the target image. We will describe the dataset collection in Section \ref{D}. In this phase, we use the trained model to extract intermediate features of the source images. We define these initial intermediate features of the encoder output as $f_{k\_i}^{res}\left( {i = 1,...,256} \right)$. $k$ represents the $kth$ input source image. In this framework, $k=1$ is for infrared images and $k=2$ is for visible images. $i$ is the total number of intermediate features of an input image. $res$ means the residual layer output. 

In addition, the output features of the first (${conv1}$) and second (${conv2}$) convolutions are defined as $f_{k\_m}^{conv1}\left( {m = 1,...,64} \right)$ and $f_{k\_n}^{conv2}\left( {n = 1,...,128} \right)$. $m$ is the total number of the features of the ${conv1}$ layer while $n$ is for the ${conv2}$ layer. As mentioned above, these features convey the texture details of the source images. We call these features as compensation features. 

Conversely, the features of the last (${dconv1}$) and the penultimate (${dconv2}$) deconvolutions are expressed as $f_{k\_m}^{dconv1}\left( {m = 1,...,64} \right)$ and $f_{k\_n}^{dconv2}\left( {n = 1,...,128} \right)$ respectively. 

However, not all information from these features is useful for reconstruction results, such as noise information. Therefore, these intermediate features need to be given different weights for refinement by using the corresponding attention maps, which will be discussed in the next section. Besides, certain details may be lost during feature extraction. Therefore, utilizing the features of each convolutional layer and passing them to the corresponding deconvolutional layer is important to compensate for the details of the reconstructed image.     

\subsection{Attention-based Feature Fusion}\label{AMG}

Recently, models based on attention mechanisms have been introduced into the training of CNN architectures. It is applied to many visual tasks, especially regions of interest (RoI) in a visual scene. The goal of infrared and visible image fusion is to retain the visual details and salient thermal radiation regions simultaneously. Therefore, motivated by previous work, we use these rough intermediate features to obtain attention maps of the source images. 

In our framework, the output of the encoder is a series of rough feature maps. Each of them depicts a special kind of information about the source images. In order to accurately reflect the salient features of the source images, we need to create attention maps from these feature maps. Each feature map has its own weight, given by the softmax operation, which calculates the probability in the channel direction.
\begin{equation}
w_i^k\left( {x,y} \right) = softmax\left( {f_{k\_i}^{res}\left( {x,y} \right)} \right),
\label{encoder}
\end{equation}
where $w_i$ is weight maps of each feature. $\left( {x,y} \right)$ denotes the same position of all feature channels. $i$ is the channel number. The softmax function can be denoted as follows:
\begin{equation}
softmax\left( {{x_1},...,{x_i}} \right) = \frac{{{e^{{x_s}}}}}{{\sum\nolimits_{j = 1}^i {{e^{{x_j}}}} }},{\rm{ }}\left( {s = 1,...,i} \right).
\end{equation}
where ${{x_i}}$ is an element of a vector sequence.

All rough feature maps are multiplied by the corresponding weights and summed to generate the attention map for the source image. The mathematic expression is as follows: 

\begin{equation}
{A^k}\left( {x,y} \right) = \sum\nolimits_{i = 1}^{256} {w_i^k\left( {x,y} \right)f_{k\_i}^{res}\left( {x,y} \right)} ,
\end{equation}
where ${A^k}$ is an attention map that reflects the activity level measurement of the source image.

According to the salient mechanism, we use the attention maps to optimize these rough features before the feature-level fusion. This process can be written as follows:
\begin{equation}
{f_i}\left( {x,y} \right) = \sum\nolimits_{j = 1}^k {w_{opt}^j\left( {x,y} \right) \times f_{k\_i}^{res}\left( {x,y} \right)},
\label{fres}
\end{equation}
with
\begin{equation}
w_{opt}^j\left( {x,y} \right) = \frac{{{A^j}\left( {x,y} \right)}}{{\sum\nolimits_{j = 1}^k {{A^j}\left( {x,y} \right)} }}.
\end{equation}
where $j$ means the $jth$ input source image. $w_{opt}^j\left( {x,y} \right)$ is the optimal weight map for the features of the $jth$ source image. ${f_i}\left( {x,y} \right)$ is the fused intermediate features which would be decoded to reconstruct the fused image.  

\subsection{Compensation Feature Fusion}\label{CFF}

For the compensation features, we can use these features to reconstruct the missing details of the convolution process in the decoder part. As each feature pixel value after compression represents a receptive field of the original image, the choose-max strategy is a better selection to merge them in an element-wise manner, which can be written as follows:
\begin{equation}
f_m^{conv1}\left( {x,y} \right) = max\left\{ {f_{1\_m}^{conv1}\left( {x,y} \right),f_{2\_m}^{conv1}\left( {x,y} \right)} \right\},
\label{first}
\end{equation}  
\begin{equation}
f_n^{conv2}\left( {x,y} \right) = max\left\{ {f_{1\_n}^{conv2}\left( {x,y} \right),f_{2\_n}^{conv2}\left( {x,y} \right)} \right\}.
\label{second}
\end{equation}
where ${f_{1\_m}^{conv1}\left( {x,y} \right)}$, ${f_{2\_m}^{conv1}\left( {x,y} \right)}$, $f_m^{conv1}\left( {x,y} \right)$ represent the infrared, visible, and fused features of the first convolution respectively. $m$ is the total number of the features of the first convolution. ${\left( {x,y} \right)}$ represents the pixel coordinate of the feature. $max\left\{  \cdot  \right\}$ is the choose-max function in an element-wise manner. The Eq.~(\ref{second}) is expressed in the same way. 

\subsection{Image Reconstruction}

Image reconstruction needs to restore the above two merged features separately. 

To begin with, the fused intermediate features ${f_i}\left( {x,y} \right),\left( {i = 1,...,128} \right)$ in Eq.~(\ref{fres}) can serve as the structural contents of the source images. We pass them to the first deconvolution of the decoder part.

In addition, the compensation features ($ f_m^{conv1}\left( {x,y} \right)$ and $ f_n^{conv2}\left( {x,y} \right)$ ) of the first and second convolutional layers can compensate the visual details of the fused image. In Section.~\ref{CFF}, we have merged them respectively in Eq.~\ref{first} and Eq.~\ref{second}. The fused compensation features are also passed to the corresponding deconvolutional layers by element-wise summation, which can be represented as follows:

\begin{equation}
f_m^1\left( {x,y} \right) = f_m^{conv1}\left( {x,y} \right) + f_m^{dconv1}\left( {x,y} \right),
\end{equation} 
\begin{equation} 
f_n^2\left( {x,y} \right) = f_n^{conv2}\left( {x,y} \right) + f_n^{dconv2}\left( {x,y} \right).
\end{equation} 
where $f_m^{dconv1}\left( {x,y} \right)$ and $f_n^{dconv2}\left( {x,y} \right)$ are the output of the first and second deconvolutional layers respectively. $f_m^1\left( {x,y} \right)$ and $f_n^2\left( {x,y} \right)$ are the input features, which are transferred into the second and the third deconvolutional layers.

Finally, the final fused image is recovered by decoding the two portions.

\section{Experimental Design}

\subsection{Dataset Preparation}\label{D}


Instead of only using visible images, we select training images from KAIST\footnote{https://soonminhwang.github.io/rgbt-ped-detection/} \cite{hwang2015multispectral} and FLIR\footnote{https://www.flir.ca/oem/adas/adas-dataset-form/} datasets, as they contain both infrared and visible versions of each image. That is more reliable than using other image datasets. The KAIST benchmark consists of 95,000 color-thermal pairs taken from a vehicle. And the FLIR dataset also provides approximately 14,452 thermal and visible image pairs for empowering the automotive community. The two datasets are recorded at 20Hz and 30Hz respectively forming into a series of video frames. Besides, the new combined dataset covers a variety of scenarios such as campuses, roads, downtown, streets, and highways. In addition, their scenes are captured during the day time and night time.     

To make the scene content of the training data more diverse, we expand the frame interval by re-sampling. The total number of the infrared (IR) and visible (RGB) images is 52,000, including 50,000 training images and 2,000 validation images. The new combined dataset is shown in the following Table.~\ref{data}.

\begin{table}[htbp]
	\caption{Dataset Collection}
	\begin{center}
		\begin{tabular}{ l l l }
			\hline
			Dataset& Training & Validation\\
			\hline
			KAIST &IR: 12,500; &IR: 500;\\
			&RGB: 12,500; &RGB: 500;\\
			\cline{2-3} 
			FLIR &IR: 12,500;&IR: 500\\
			& RGB: 12,500;& RGB: 500;\\
			\hline
			Total Number&50,000 & 2,000 \\
			\hline					
		\end{tabular}
		\label{data}
	\end{center}
\end{table}

All of these images are resized to $256 \times 256$ pixels and converted to gray-scale images, which are further normalized to [0, 1] interval. 

\subsection{Training Details} \label{T}

The training stage combines a symmetric encoder-decoder and residual block (SEDR) into a deep network. The SEDR has six layers, including three convolutional layers and three deconvolutional layers, between which there is a residual block. In the encoder part, the output size is half the input size and the number of features is twice that of the previous layers. The decoder part is just the opposite with the corresponding convolutional layers. The intermediate features can be fully reused through the residual block. The output intermediate features from the residual block are $64 \times 64 \times 256$. 

We train the SEDR network on the prepared dataset (see Table.~\ref{data}). The batch number and epoch number are set to 2 and 50 respectively. The learning rate is set to $1 \times {10^{ - 4}}$. The same with Li et al. \cite{li2018densefuse}, we still use the pixel-loss and SSIM-loss as the total loss function. These two loss functions can constraint the reconstructed pixel error and edge error respectively. The mathematical expression of the total loss is as follows: 
\begin{equation}
{T_{loss}} = {P_{loss}} +  SSI{M_{loss}},
\label{loss}
\end{equation}
where ${T_{loss}}$, ${P_{loss}}$, and $SSI{M_{loss}}$ represent the total loss, pixel loss, and SSIM loss respectively. In addition, the SSIM loss is generated by taking 1 and subtracting the structural similarity value computed in \cite{wang2004image}, which can be written as follows: 
\begin{equation}
SSI{M_{loss}} = 1 - SSIM\left( {out,in} \right).
\label{SS}
\end{equation}
\begin{equation}
{P_{loss}} = \sqrt {\frac{1}{{MN}}\sum\limits_{x \in M,y \in N} {{{\left( {out\left( {x,y} \right) - in\left( {x,y} \right)} \right)}^2}} }.
\label{ploss}
\end{equation}
where ${out}$ and ${in}$ mean the reconstructed data and input training data, respectively. $SSIM\left(  \cdot  \right)$ indicates the SSIM function. In Eq.~(\ref{ploss}), $M$ and $N$ is the size of an image. ${\left( {x,y} \right)}$ is the pixel location.       

Our framework is implemented with NVIDIA GTX 1070Ti (GPU), 32GB RAM (Memory), and Intel Core i5-8500 (CPU). The network architecture is programmed on the Tensorflow.

\begin{figure}[h]
	\centering
	\includegraphics[width=9cm]{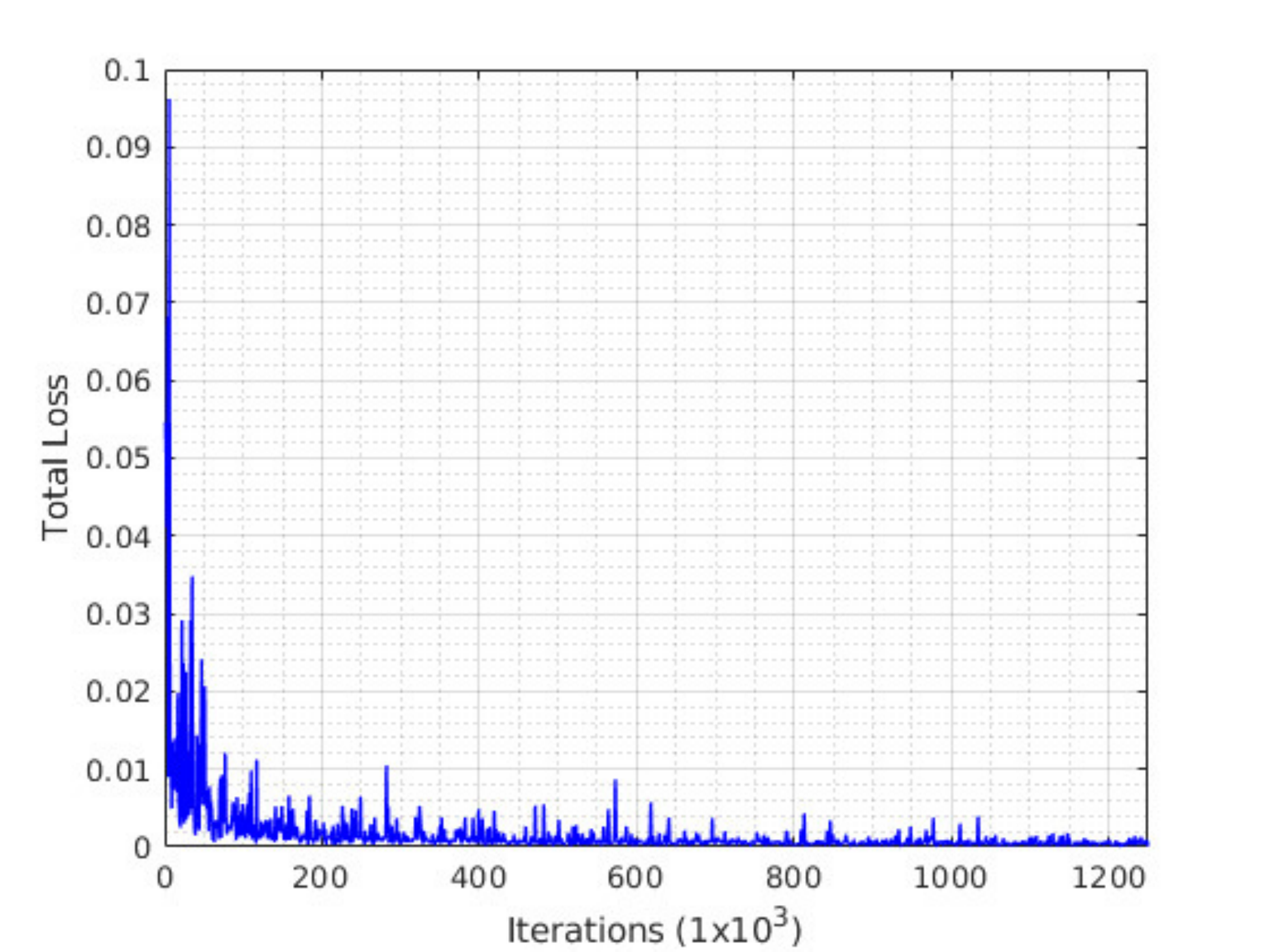}
	\caption{The total loss curve in the training stage.}
	\label{totalloss}
\end{figure}

Fig.~\ref{totalloss} shows the training total loss curve on the new dataset. Every 1000 iterations output a total loss value. In this work, each epoch needs $25\times {10^{ 3}}$ iterations. From the curve, we can see that the total loss value tends to be stable around $1200 \times {10^{ 3}}$ iterations (or 48 epochs). It demonstrates that the trained model has reached optimally. 

\begin{figure}[h]
	\centering
	\includegraphics[width=9cm]{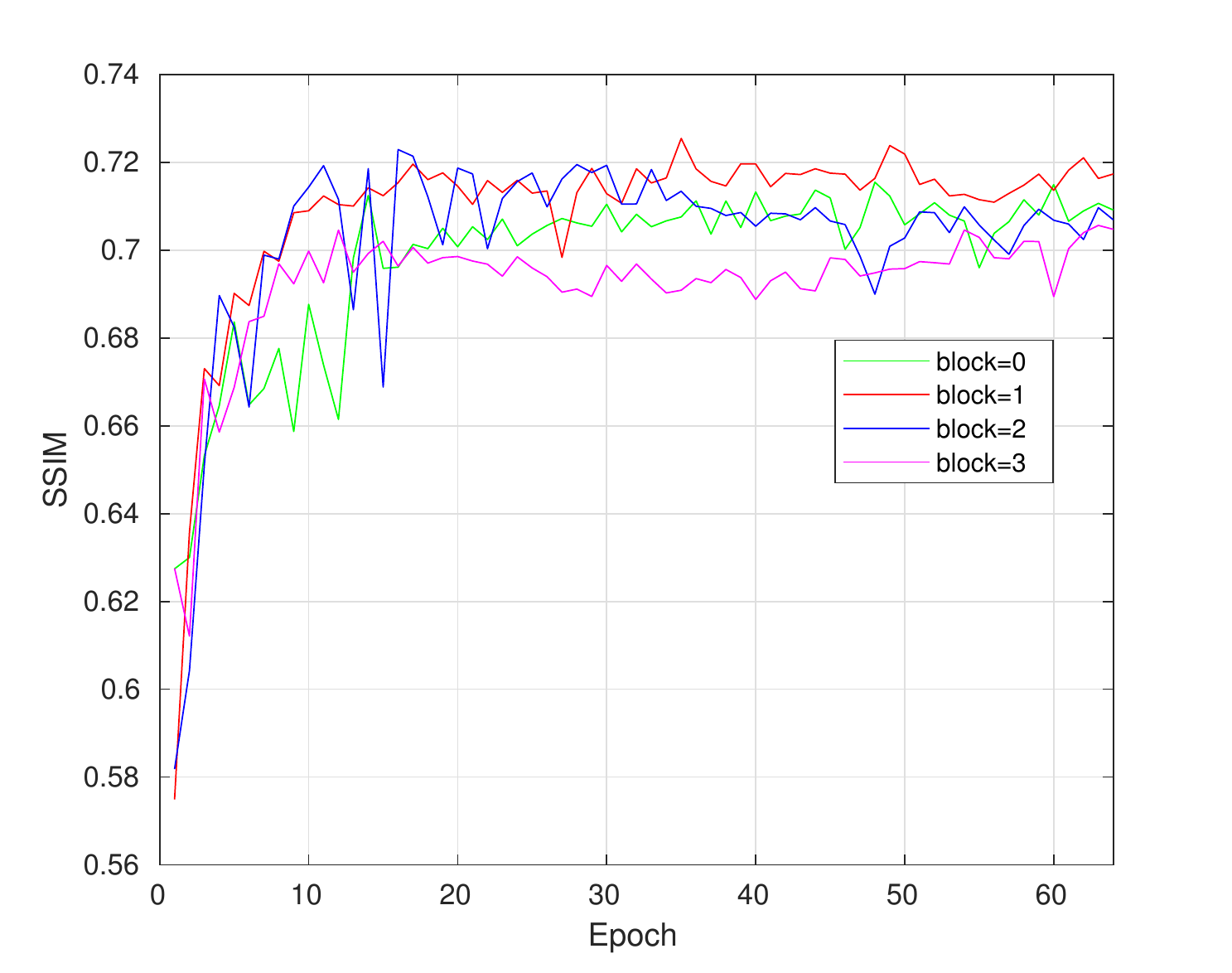}
	\caption{The relationship curve between the number of residual blocks and corresponding SSIM values.}
	\label{block}
\end{figure}

To investigate the effect of the number of residual blocks on the fusion results, we choose the SSIM metric as the performance measurement, as shown in Fig.~\ref{block}. It can be seen that the proposed framework achieves the best performance with one residual block (see red curve). When the number of the residual blocks is increased, the SSIM value will decrease. In addition, more residual blocks result in time-consuming during the training stage. Therefore, in this work, we design the framework using one residual block. 

\begin{figure*}[h]
	\centering
	\includegraphics[width=19cm]{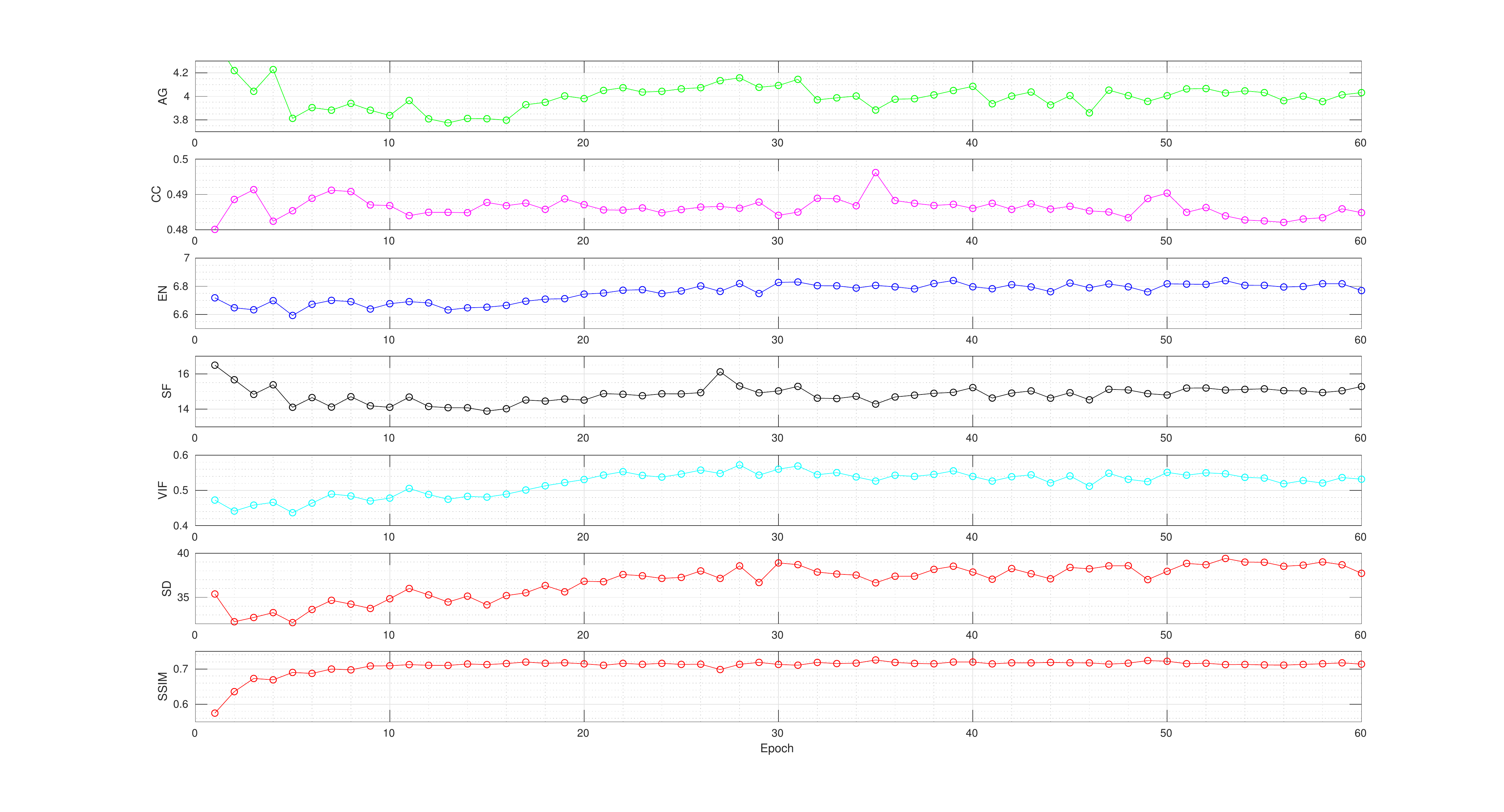}
	\caption{Metric change curves with the increasing of the training epoch number.}
	\label{los}
\end{figure*}

\subsection{Experimental Setting}

\begin{figure*}[h]
	\centering
	\includegraphics[width=17cm]{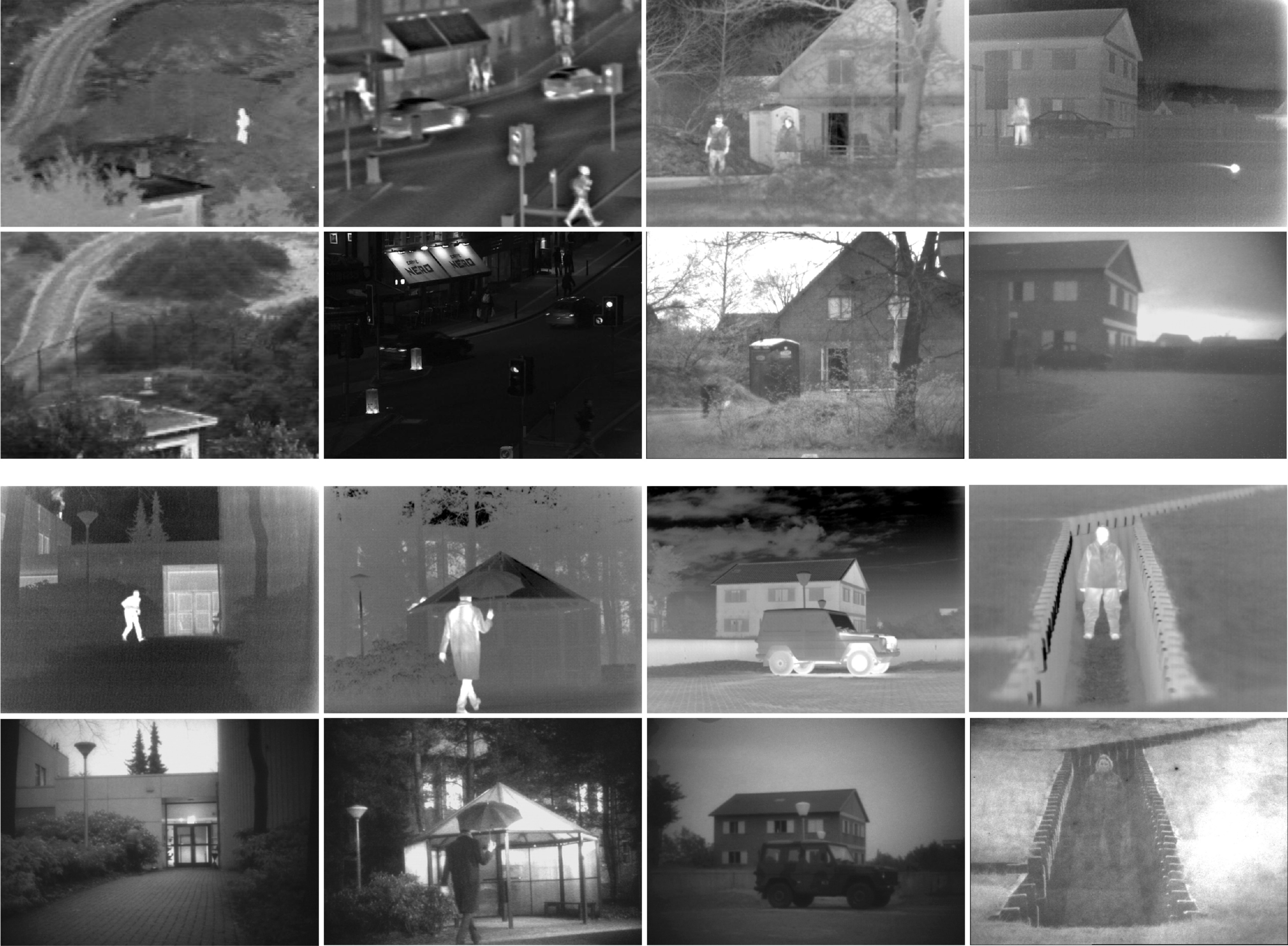}
	\caption{A portion of source images from ``$TNO~image~fusion~dataset$": the first and third rows are infrared images, the second and fourth rows are corresponding visible images.}
	\label{source}
\end{figure*}

In this experiment, the test source images are derived from the $TNO~image~fusion~ dataset$\footnote{https://figshare.com/articles/TNO\_Image\_Fusion\_Dataset/1008029}, from which we select 20 pairs of infrared and visible images containing different scenes. Before implementing image fusion, all the source image pairs should be strictly aligned. A portion of source images in this paper is shown in Fig.~\ref{source}. 

Our comparative experiment will be implemented on five existing methods, including CNN-based fusion \cite{liu2017multi}, Deep-Fuse \cite{ram2017deepfuse}, DenseFuse \cite{li2018densefuse}, FusionGAN \cite{ma2019fusiongan}, and GFF \cite{li2013image}. The fusion results are shown in Figs.~\ref{test1}-\ref{test2}, which will be analyzed in Section~\ref{R}. Parameters for all comparable fusion methods are strictly in accordance with the settings given by authors.

For validating the performance of the fusion results, apart from subjective and visual comparison, we also use seven quantitative fusion metrics to confirm the effectiveness of our proposed method. These fusion metrics are described in the following Section.  

\subsection{Fusion Evaluation Metrics}
To evaluate the quality of image fusion, it is not enough to solely rely on subjective evaluation, because some fusion methods produce similar visual results. In this paper, except for the visual evaluation, we also use seven quantitative metrics to objectively evaluate the performance of the fused results. These metrics are briefly listed as follows:
\begin{itemize}
	\item Entropy ($EN$) \cite{liu2015general}. 
	
	$EN$ of an image reflects the total information involved in a synthetic image in terms of information theory. The larger the $EN$ value, the better performance the fusion result. However, this metric alone cannot determine the overall quality of the fused image. 
	
	\item Spatial frequency ($SF$) \cite{zheng2007new}.
	
	$SF$ measures the overall activity level of an image. $SF$ combines four direction spatial frequencies, including the row, column, main diagonal, and secondary diagonal.  It can indicate the structural and textural information of the fused image. The $SF$ metric shows good fusion result at a high value.  
	
%
	
	\item Standard deviation ($SD$). 
	
	$SD$ employs a statistical approach to calculate the distance between each individual pixel and the mean in an image. It reflects the dispersion of image pixel value and the mean. The larger the standard deviation, the better the image quality. In addition, a large $SD$ value has high spatial contrast in the fused image.
	
	\item Average gradient ($AG$). 
	
	$AG$ is a definition of the sharpness of an image, reflecting the ability of the image to express contrast. Specifically, it reflects the change in the tiny details of the image, as well as the ratio of contrast and relative sharpness in the multi-dimensional direction of the image. The larger the $AG$ value is, the better the fused performance has.  
	
	\item Correlation coefficient ($CC$) \cite{deshmukh2010image}. 
	
	$CC$ represents the degree of linear correlation of the fused image and source images. A higher $CC$ value signifies that the fused image is more similar to the source images. It represents good fusion performance when we obtain a larger value.
	     
%
	
	\item Structure similarity ($SSIM$) \cite{wang2004image}.
	
	$SSIM$ is a combination of correlation, luminance, and contrast distortion. This metric is consistent with human visual sensitivity in terms of structure loss and distortion. The $SSIM$ value ranges from -1 to 1, in which -1 and 1 indicate converse and same structure with the reference image respectively, whereas 0 represents no relationship with the reference image. A high positive value means a good fusion quality.   
	
	\item Visual information fidelity ($VIF$) \cite{han2013new}.
	
	$VIF$ mainly computes the information fidelity of the fused image. This metric leverages different models, such as the human visual system (HVS) model, the natural scene statistics (NSS) model, and the distorted model, to extract mutual information from each block and sub-band. The larger the $VIF$ value, the excellent the fusion result are. 
	
\end{itemize}

Fig.~\ref{los} shows the relationship between the training epoch number and fusion metrics on the test images. Combining the training total loss curve in Fig.~\ref{totalloss}, after 50 epochs, almost all fusion metrics fluctuate within a small range except for $CC$. Overall, our proposed fusion method performs well in terms of quantitative evaluation. Therefore, setting the epoch number to 50 is reasonable in our training stage. 

\subsection{Results Analysis}\label{R}

In this section, we use both subjective visual evaluation and objective quantitative evaluation to analyze the fused results for six existing image fusion methods. Two infrared and visible image pairs are selected for experimental comparison because of space limitations. The remaining image pairs have similar effects.

\begin{figure*}[h]
	\centering
	\includegraphics[width=17cm]{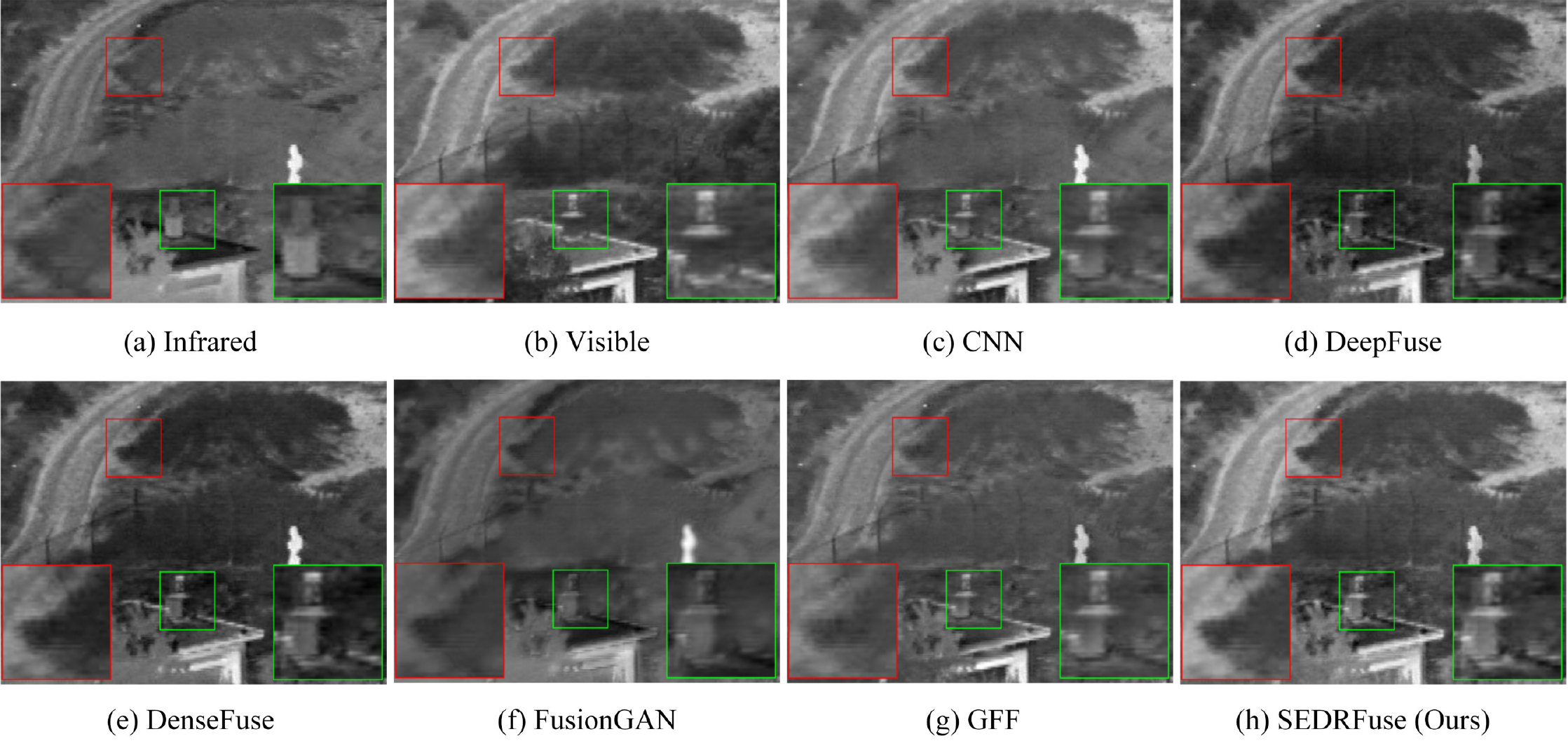}
	\caption{Fusion results on ``$Nato\_camp$". Top row-from left to right: infrared image, visible image, results of GFF-, CNN- based methods; bottom row-from left to right: results of DeepFuse-, DenseFuse-, FusionGAN- based methods, and proposed fusion method.}
	\label{test1}
\end{figure*}

\begin{figure*}[h]
	\centering
	\includegraphics[width=17cm]{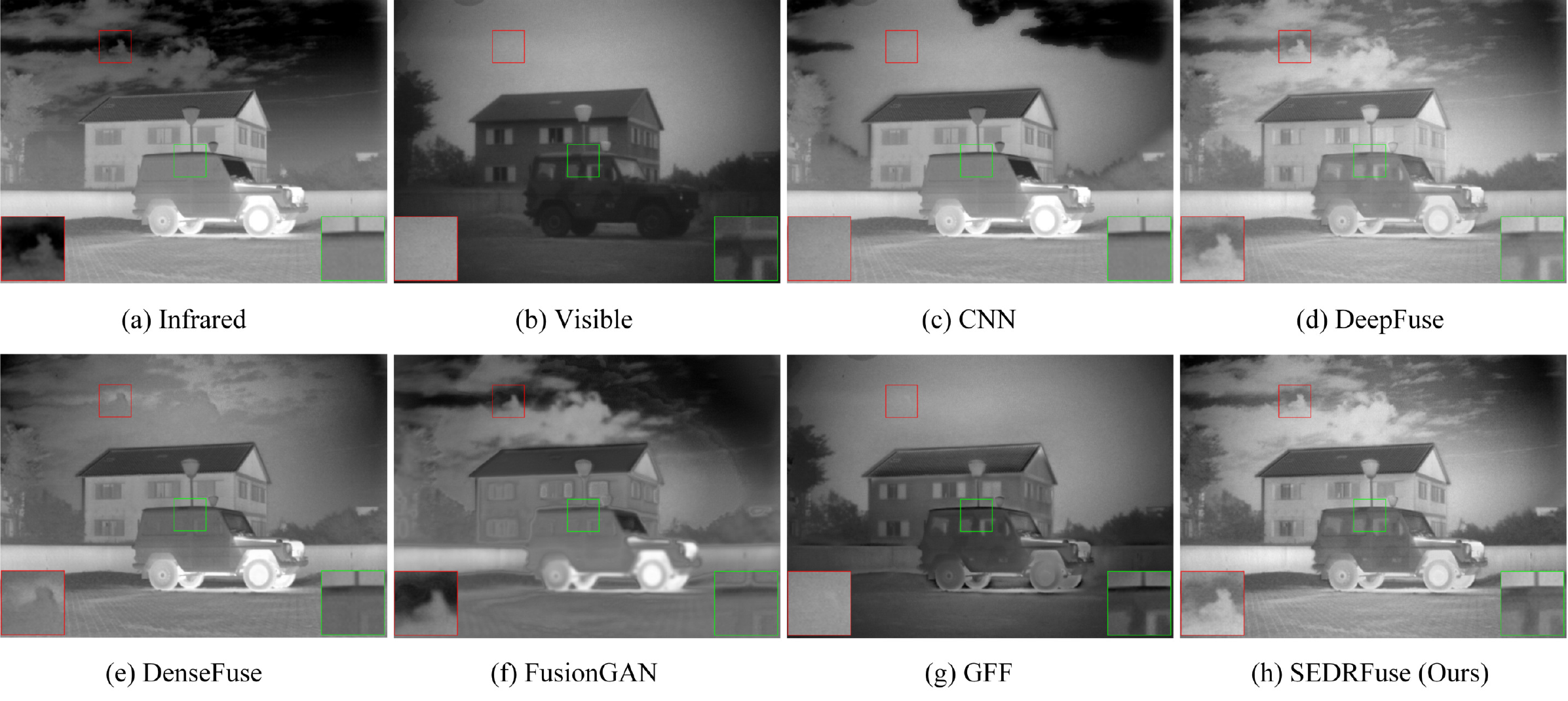}
	\caption{Fusion results on ``$Marne\_04$". Top row-from left to right: infrared image, visible image, results of GFF-, CNN- based methods; bottom row-from left to right: results of DeepFuse-, DenseFuse-, FusionGAN- based methods, and proposed fusion method.}
	\label{test2}
\end{figure*}

From the visual view, in Figs.~\ref{test1}-\ref{test2}, the results by six fusion methods (CNN-, DeepFuse-, DenseFuse-, FusionGAN-, GFF- based methods, and the proposed method) are listed as (c) to (h). Figures~(a) and (b) represent infrared and visible images, respectively. Results generated by the CNN-based method shows obvious block effects (see Fig.~\ref{test1}(c) and Fig.~\ref{test2}(c)). This method is limited to the fusion of multi-focus images and has no generalization on other types of multi-modality image fusion. The DenseFuse-based method does not perform well. For example, the brightness of the road and the roof are dim compared to the original source images (see Fig.~\ref{test1}(e)). In addition, the shape of the cloud is very unclear in Fig.~\ref{test2}(e). The fused results obtained by FusionGAN-based method cannot reflect the visible detail clearly (see Fig.~\ref{test1}(f) and Fig.~\ref{test2}(f)). The GFF-based method struggles to pass the infrared information to the fused image and has low brightness. The close-up areas of Fig.~\ref{test1}(g) have low brightness and contrast. Similarly, the close-up areas of Fig.~\ref{test2}(g) show that the cloud information cannot be transferred to the fused image. Although the DeepFuse-based (see Fig.~\ref{test1}(d) and Fig.~\ref{test2}(d)) and the proposed method (see Fig.~\ref{test1}(h) and Fig.~\ref{test2}(h)) have very close visual performance, the results by the proposed method are more natural and show clear structural features. Therefore, in terms of infrared and visible image fusion, the proposed method is superior to the existing methods in visual evaluation.     

Table.~\ref{tab1} shows the objective evaluation of the six image fusion methods using the two selected image pairs. The proposed method always has larger values in terms of $AG$, $CC$, $EN$, $SF$, and $VIF$ metrics. The $SSIM$ values are not the best in $Nato\_camp$ and $Marne\_04$ image fusions using our method. It may be the effect of other factors on outcome of the fusion, such as the content of the scene. However, in general, our fusion results are not poor in terms of $SSIM$ metric. Consistent performance of the subjective and objective evaluations strongly demonstrates that the proposed framework is more efficient than the existing fusion methods.

\begin{table*}[ht]
	\caption{Objective quality of the selected three image pair fusion by different methods}
	\label{tab1}
	\begin{center}
		\begin{tabular}{ l l l l l l l l }
			\hline
			Source images       &Methods      &CNN              &DeepFuse        &DenseFuse     &FusionGAN     &GFF    &SEDRFuse (Ours)   \\
			\hline
			$Nato\_camp$
			& $AG$             &3.7923	      &3.9980	         &4.1615	     &2.4636	    &3.6798	         &\textbf{4.4323}   \\
			& $CC$             &0.5763	      &0.6773	         &0.6587	     &0.5290	    &0.6238	         &\textbf{0.6699}   \\
			& $EN$             &6.8151        &6.7326            &6.9035         &6.5267        &6.3626          &\textbf{6.9828}  \\
			& $SF$             &11.1157	      &14.7569	         &16.0202	     &9.4881	    &14.7782	     &\textbf{16.2545}  \\
			& $SSIM$           &0.7118	      &0.7354	         &0.7056	     &0.6987	    &\textbf{0.7408} &\textcolor{red}{0.7340}  \\
			& $VIF$            &0.2671	      &0.5697	         &0.4727	     &0.1413	    &0.2610	         &\textbf{0.6121}   \\
			& $SD$             &31.2828	      &34.4454	         &36.7658	     &24.9111       &25.8975	     &\textbf{37.5175}   \\
			
			\hline
			$Marne\_04$  
			& $AG$             &2.7619	      &3.0682	         &2.6843	     &2.2036	    &2.4038	         &\textbf{3.5356}    \\
			& $CC$             &0.2835	      &0.3777	         &0.2753	     &0.2700	    &0.2228	         &\textbf{0.3903}    \\
			& $EN$             &7.2929        &7.2917            &7.0827         &7.1004        &7.0134          &\textbf{7.3159}    \\
			& $SF$             &10.8251	      &10.9930	         &10.4965	     &8.2041	    &9.6045	         &\textbf{13.0429}   \\
			& $SSIM$           &0.6828	      &0.6718	         &\textbf{0.7114} &0.6608       &0.6949	         &0.6650             \\
			& $VIF$            &0.3652	      &0.7078	         &0.3493	     &0.4229	    &0.1132	         &\textbf{0.8277}    \\
			& $SD$      &\textbf{49.0840}	  &40.1985	     &36.6558	 &42.6331       &36.2172	     &\textcolor{red}{44.7425}   \\
			\hline
		\end{tabular}
	\end{center}
\end{table*}

The proposed method is not limited to the current few image pairs, as it is validated across the entire test dataset to obtain the average evaluation values in Table.~\ref{tab2}. We can see that the proposed fusion method outperforms previous fusion methods in all metrics except the $SSIM$ and $SD$, which it places third and second respectively. Our proposed method also specifically excels at the $AG$ and $SF$ metrics. In general, these excellent results exhibit that our fusion method can preserve the structural components of the source images, and transfer the visual details to the fused results simultaneously.

\begin{table*}[ht]
	\caption{Average performance of different image fusion methods on the TNO dataset.}
	\label{tab2}
	\begin{center}
		\begin{tabular}{ l l l l l l l l}
			\hline
			Methods       &$AG$ &$CC$ &$EN$  &$SF$  &$SSIM$  &$VIF$   & $SD$            \\
			\hline

			CNN             & 3.6984   &0.3784  &6.7758  &14.2968   &0.7126  &0.2382   &\textbf{40.4855 } \\
			
			DeepFuse     & 3.5302    &0.4907	  &6.6525     &12.8103 &\textbf{0.7308} &0.5281  &33.0762   \\
			
			DenseFuse    & 3.4587      &0.4454	 &6.8080   &13.2573	 & 0.7196 &0.3596  &37.6247  \\
			
			FusionGAN    & 2.1586       &0.4194	      &6.3275		&8.0845	   & 0.6575  &0.1827   &25.7923 \\
			
			GFF               & 3.8225    &0.3420 &6.7877	  &14.5258	 &0.7243 &0.2473   &37.0476  \\
			
			SEDRFuse (Ours)           & \textbf{4.0054}      &\textbf{0.4910}	  &\textbf{6.8179}	 &\textbf{14.7979} & \textcolor{red}{0.7219}   &\textbf{0.5511} &\textcolor{red}{37.9516} \\
			
			\hline
		\end{tabular}
	\end{center}
\end{table*}

\begin{figure*}[h]
	\centering
	\includegraphics[width=17cm]{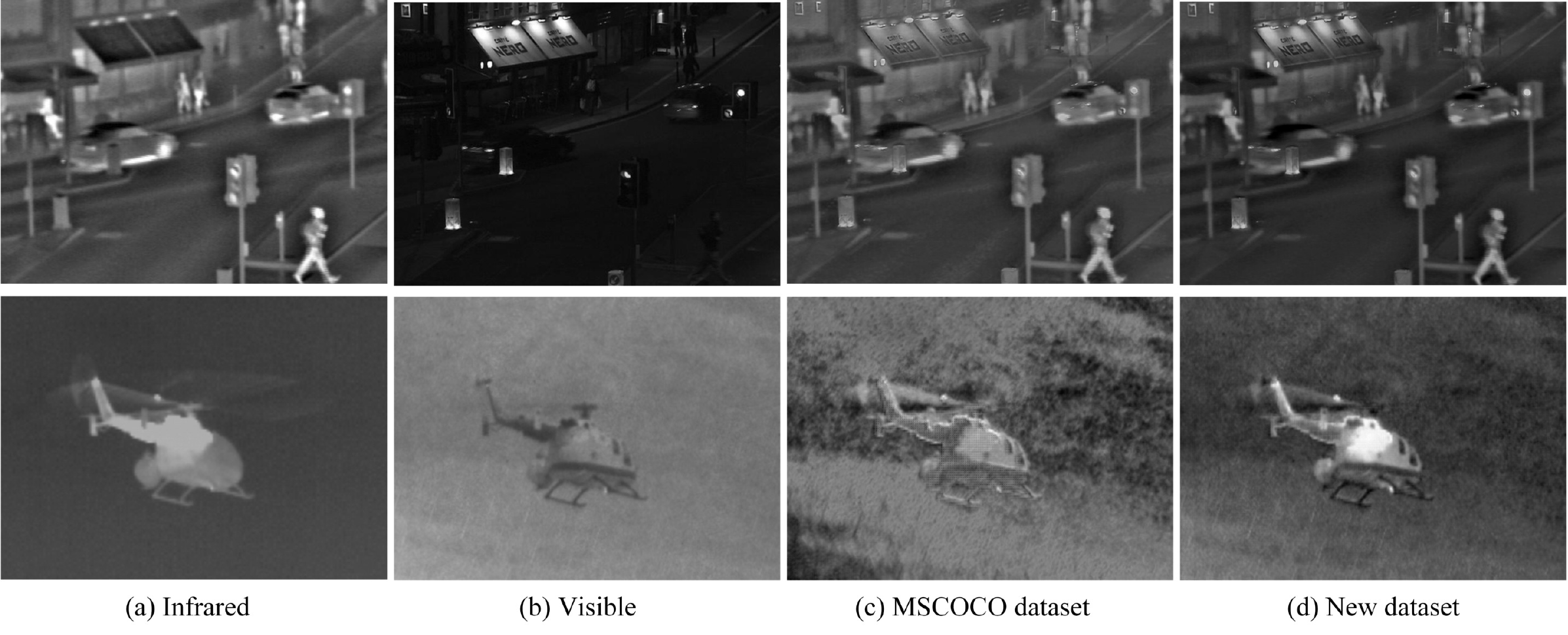}
	\caption{Fusion results by using different training datasets.(a) infrared images; (b) visible images; (c) results by using MSCOCO dataset; (d) results by using New dataset.}
	\label{compare}
\end{figure*}

We also verified the influence of different training datasets on fusion results. Except for the training dataset, the comparison experiments adopt the same parameter settings and pre-processing way. We use the MSCOCO dataset\footnote{http://cocodataset.org/download/} as a training dataset to demonstrate that only visible images cannot compare with the specific datasets (FLIR and KAIST) in terms of infrared and visible image fusion. Fig.~\ref{compare} shows the fusion results by using different training datasets. It can be seen that the MSCOCO dataset displays poor performance (see Fig.~\ref{compare}(c)). In contrast, fusion results from the new training dataset tend to be more natural and better (see Fig.~\ref{compare}(d)). Table.~\ref{tab} gives average fusion performance on the test dataset by using different training datasets. It can be seen that the new combined datasets can achieve better objective results than the MSCOCO dataset. Therefore, data selection is very important for training networks to conduct different fusion tasks.

\begin{table*}[ht]
	\caption{Average performance on the TNO dataset by different training datasets.}
	\label{tab}
	\begin{center}
		\begin{tabular}{ l l l l l l l l}
			\hline
			Datasets       &$AG$ &$CC$ &$EN$  &$SF$  &$SSIM$  &$VIF$   & $SD$            \\
			\hline						
			MSCOCO               & 3.9384    &0.4532 &6.7586	  &14.4086	 &0.6914 &0.4831   &\textbf{38.9456}  \\
			
			FLIR and KAIST           & \textbf{4.0054}      &\textbf{0.4910}	  &\textbf{6.8179}	 &\textbf{14.7979} & \textbf{0.7219}   &\textbf{0.5511} &37.9516 \\
			
			\hline
		\end{tabular}
	\end{center}
\end{table*}

Table.~\ref{tab3} compares the average running time of different fusion methods to fuse one infrared and visible image pair. This shows that the proposed method is adequately fast.       

\begin{table*}[ht]
	\caption{Average running time of one image pair fusion by different methods}
	\label{tab3}
	\begin{center}
		\begin{tabular}{ l l l l l l l }
			\hline
			Methods      &GFF &CNN     &DeepFuse    &DenseFuse    &FusionGAN       &SEDRFuse (Ours)       \\
			\hline			
			Time (seconds per pic)      &0.258 &86.069      &0.381	      &2.333	 &1.637	      &1.635   \\	
			\hline
		\end{tabular}
	\end{center}
\end{table*}

\subsection{Other Types of Image Fusion}

\begin{figure*}[h]
	\centering
	\includegraphics[width=17cm]{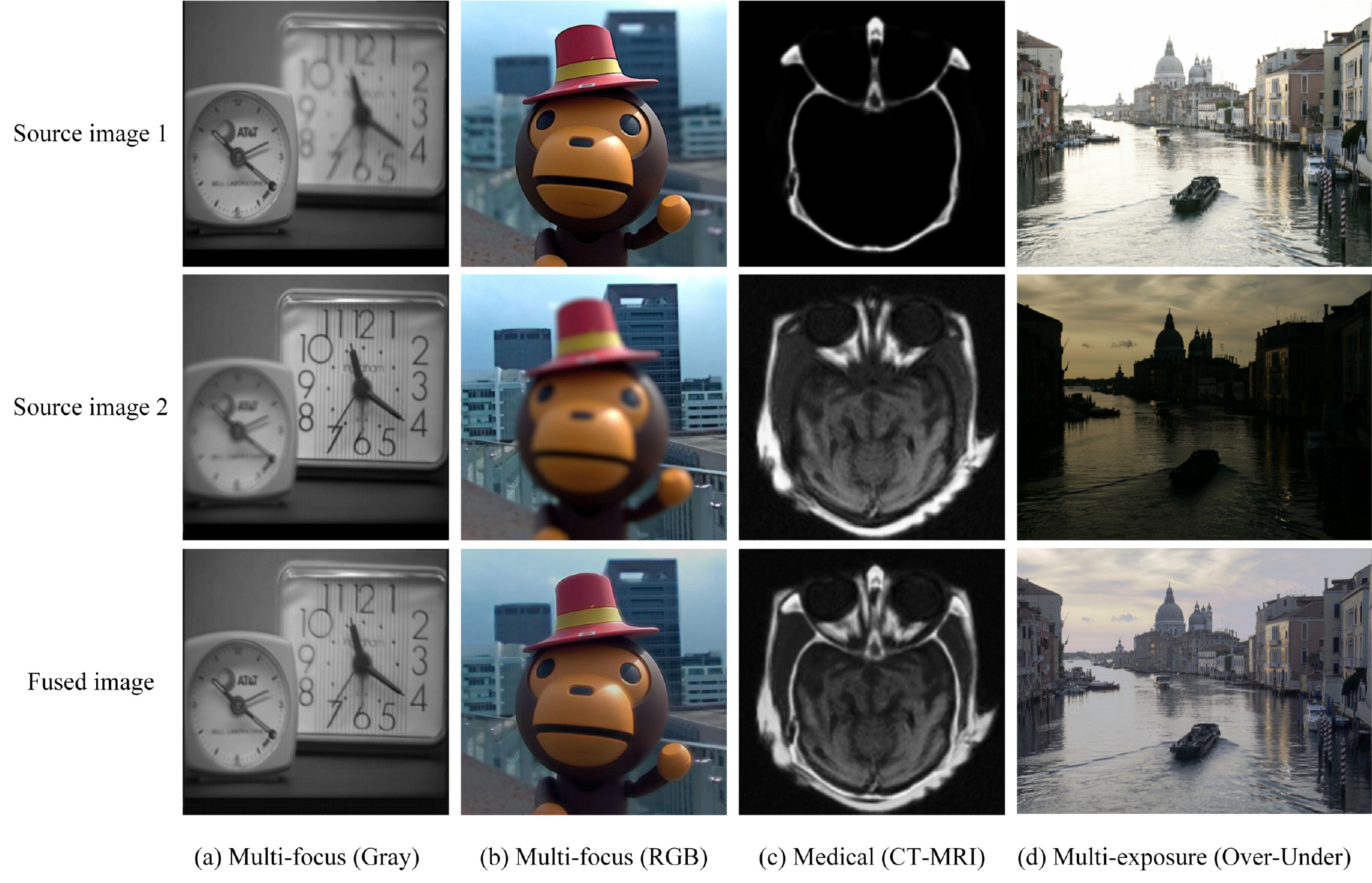}
	\caption{Fusion results on other types of images.}
	\label{other}
\end{figure*}

To demonstrate the generalization capability of the proposed method, we attempt to extend its applications on other types of multi-modality images, including multi-focus images (gray scale and color scale, see Fig.~\ref{other}(a)-(b)), medical images (CT and MRI, see Fig.~\ref{other}(c)), and multi-exposure images (over-exposure and under-exposure, see Fig.~\ref{other}(d)). Although the model has not been trained in the relevant datasets, the fusion results still perform good visual effects. It can be seen that the structures and details of the source images are transferred well into the fused image. 

Therefore, it demonstrates that our proposed method is also applicable to image fusion of other modalities, and that it can achieve good performance.

\section{Conclusion}
The infrared and visible image fusion is a well-studied problem in image processing. This article aims to explore a two-stage feature-level fusion method for infrared and visible images. In addition, inspired by the restoration capability of the encoder-decoder network, we design a symmetric network combining the residual block (SEDR) as a fixed feature extractor. The training purpose of this SEDR network is to reduce the error of the input and output data. Only if the training loss reaches a stable value, the intermediate output features are representative. In our framework, the training datasets come from KAIST and FLIR published online. We extract the intermediate features of the source images and generate two attention maps, which are in turn used to fuse the intermediate features. The features generated by the first two convolutional layers are also utilized to fuse for compensating the detail loss in down-sampling. Finally, the fused intermediate features and the fused compensation features are fed back into the decoder and the corresponding deconvolutional layers respectively, to reconstruct the fused image.

Our fusion framework focuses on combining infrared and visible images on a feature-level. Using our framework, the time-consuming problem of the traditional pixel-based image fusion methods can be significantly reduced. This feature-level image fusion can avoid generating redundant information in the fused image to some extent. Overall, the proposed fusion method achieves better results than the-state-of-the-art methods in terms of visual and objective evaluations. Future study will focus on feature-level fusion, and improving the use of intermediate features to fuse other modality images.  


%

%

\section*{Acknowledgment}
This research is sponsored by National Natural Science Foundation of China (No. 61701327, No.
61711540303, and No. 61601266), Science Foundation of Sichuan Science and Technology Department (No. 2018GZ0178), also is supported by Graduate Student’s Research and Innovation Fund of Sichuan University (Grant No. 2018YJSY058), the Priority Academic Program Development of Jiangsu Higher Education Institutions (PAPD) Fund, Jiangsu Collaborative Innovation Center on Atmospheric Environment and Equipment Technology (CICAEET) Fund. The authors thank the financial support by China Scholarship Council (Grant No. 201806240047), and also thank Dr. Zheng Liu for his helpful guidance.

\ifCLASSOPTIONcaptionsoff
  \newpage
\fi




\bibliographystyle{IEEEtran}
\bibliography{fusion}

%



%
\begin{IEEEbiography}[{\includegraphics[width=1in,height=1.25in,clip,keepaspectratio]{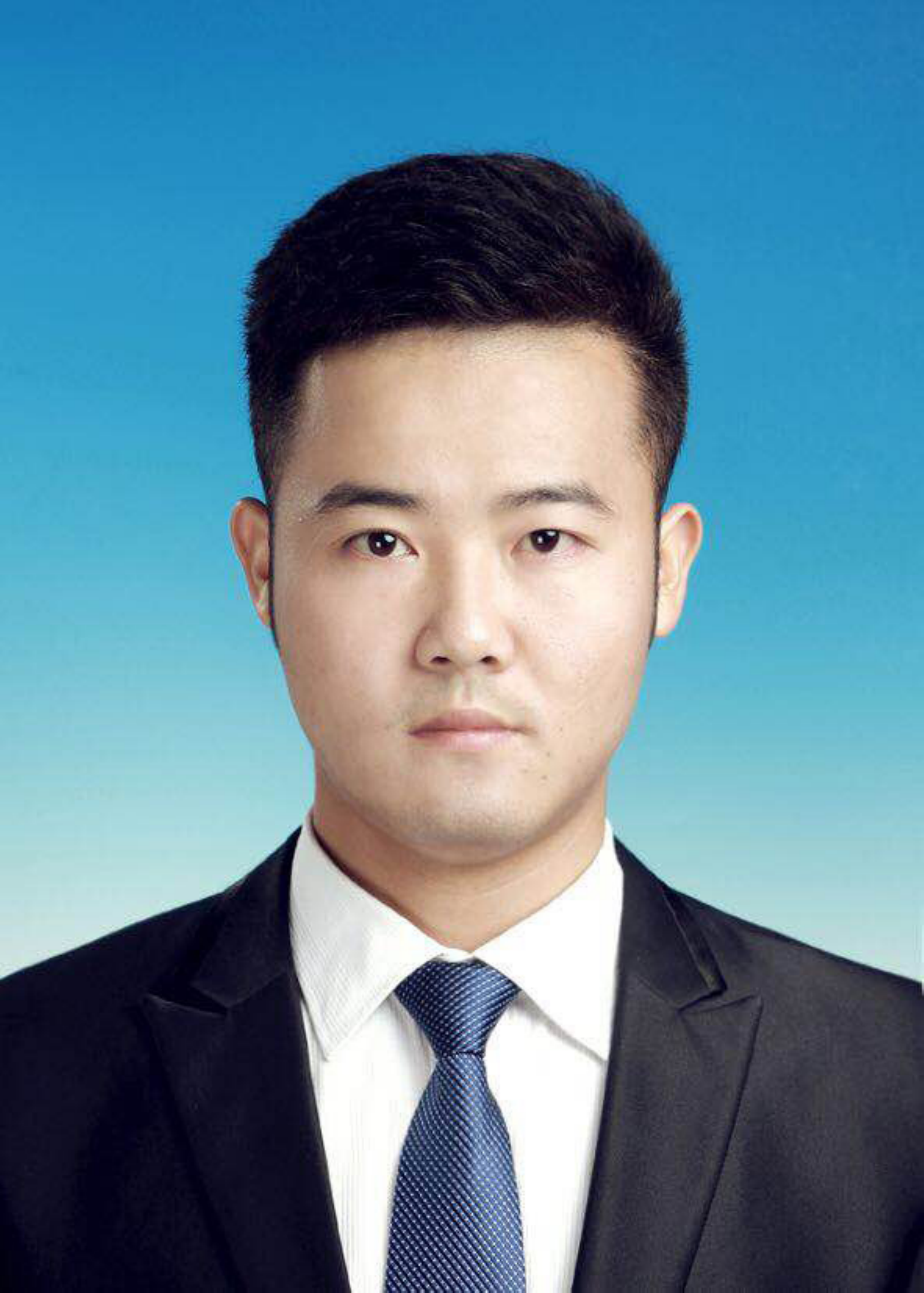}}]{Lihua Jian}
received his M.S. degree of Instrument Science and Technology from University of Electronic Science and Technology of China. He is currently pursuing Ph.D. degree in College of Electronics and Information Engineering, Sichuan University. Now, he is a visiting Ph.D. student with the School of Engineering, University of British Columbia, BC, Canada. His research interests are image processing and computer vision.
\end{IEEEbiography}

\begin{IEEEbiography}[{\includegraphics[width=1in,height=1.25in,clip,keepaspectratio]{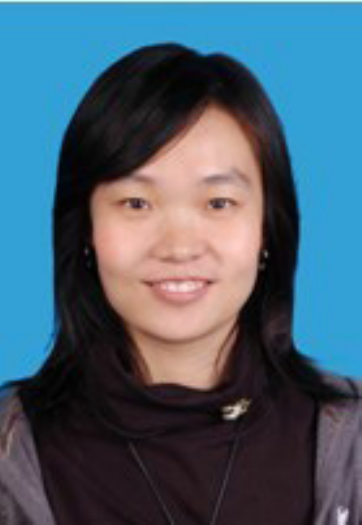}}]{Xiaomin Yang}
is currently a Professor in College of Electronics and Information Engineering, Sichuan University. She received her B.S. degree from Sichuan University, and received her Ph.D. degree in communication and information system from Sichuan University. She worked in University of Adelaide as a post doctorate for one year. Her research interests are image processing and pattern recognition.
\end{IEEEbiography}

\begin{IEEEbiography}[{\includegraphics[width=1in,height=1.25in,clip,keepaspectratio]{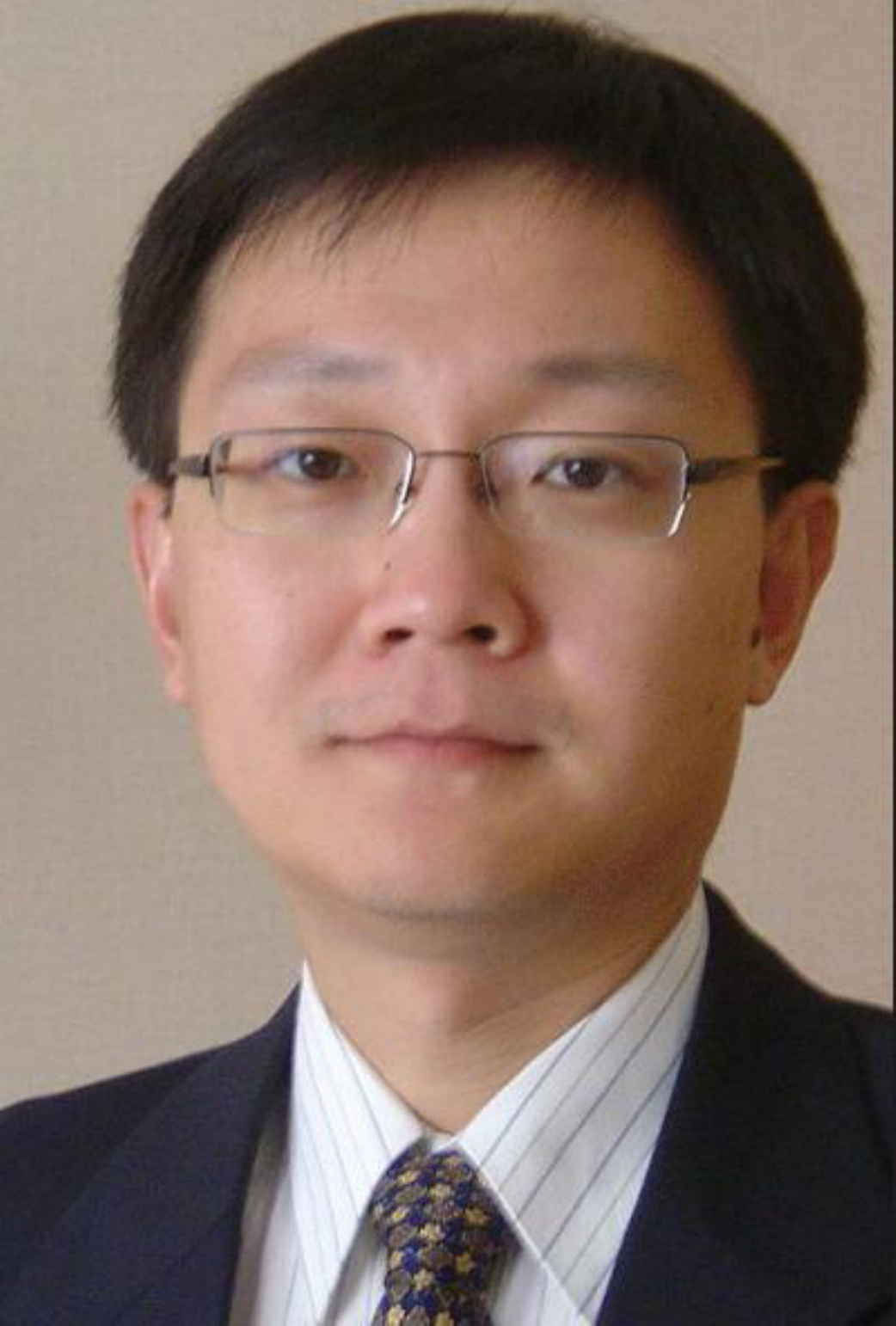}}]{Zheng Liu}
received the Ph.D. degree in engineering from Kyoto University, Kyoto, Japan, in 2000, and the second Ph.D. degree from the University of Ottawa, Canada, in 2007. From 2000 to 2001, he was a Research Fellow with Nanyang Technological University, Singapore. He then, joined the Institute for Aerospace Research (IAR), National Research Council Canada, Ottawa, ON, Canada, as a Governmental Laboratory Visiting Fellow nominated by NSERC. After being with IAR for five years, he transferred to the NRC Institute for Research in Construction, where he was a Research Officer. From 2012 to 2015, he was a Full Professor with the Toyota Technological Institute, Nagoya, Japan. 

He is currently with the School of Engineering, The University of British Columbia, Okanagan. His research interests include image/data fusion, computer vision, pattern recognition, senor/sensor network, condition-based maintenance, and non-destructive inspection and evaluation. He is a member of SPIE. He is chairing the IEEE IMS technical committee on industrial inspection (TC-36). He holds a Professional Engineer license in British Columbia and Ontario. He serves on the editorial board for journals the IEEE Transactions on Instrumentation and Measurement, the IEEE Instrumentation and Measurement Magazine, the IEEE Journal of RFID, Information Fusion, Machine Vision and Applications, and Intelligent Industrial Systems.
\end{IEEEbiography}

\begin{IEEEbiography}[{\includegraphics[width=1in,height=1.25in,clip,keepaspectratio]{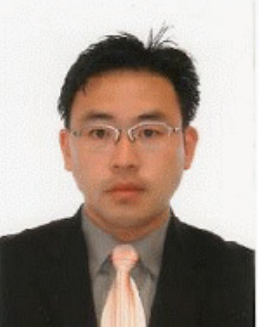}}]{Gwanggil Jeon}
	received the B.S., M.S., and Ph.D. degrees from Hanyang University, in 2003, 2005, and 2008, respectively. From 2009 to 2011, he was a Postdoctoral Fellow with the University of Ottawa, Ottawa, ON, Canada. From 2011 to 2012, he was an Assistant Professor with Niigata University. He is currently a Professor with Xidian University, Xi’an, China, and Incheon National University, Incheon, South Korea. His research interests include image processing, particularly image compression, motion estimation, demosaicking, and image enhancement as well as computational intelligence such as fuzzy and rough sets theories.
\end{IEEEbiography}

\begin{IEEEbiography}[{\includegraphics[width=1in,height=1.25in,clip,keepaspectratio]{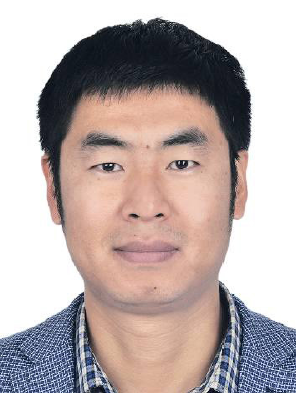}}]{Mingliang Gao}
received the Ph.D. degree in communication and information systems from Sichuan University, Chengdu, China, in 2013. From 2018 to 2019, he was a Visiting Scholar with the School of Engineering, The University of British Columbia, Okanagan. He is currently an Associate Professor with the School of Electrical and Electronic Engineering, Shandong University of Technology, Zibo, China. His main research interests include image fusion and deep learning.
\end{IEEEbiography}

\begin{IEEEbiography}[{\includegraphics[width=1in,height=1.25in,clip,keepaspectratio]{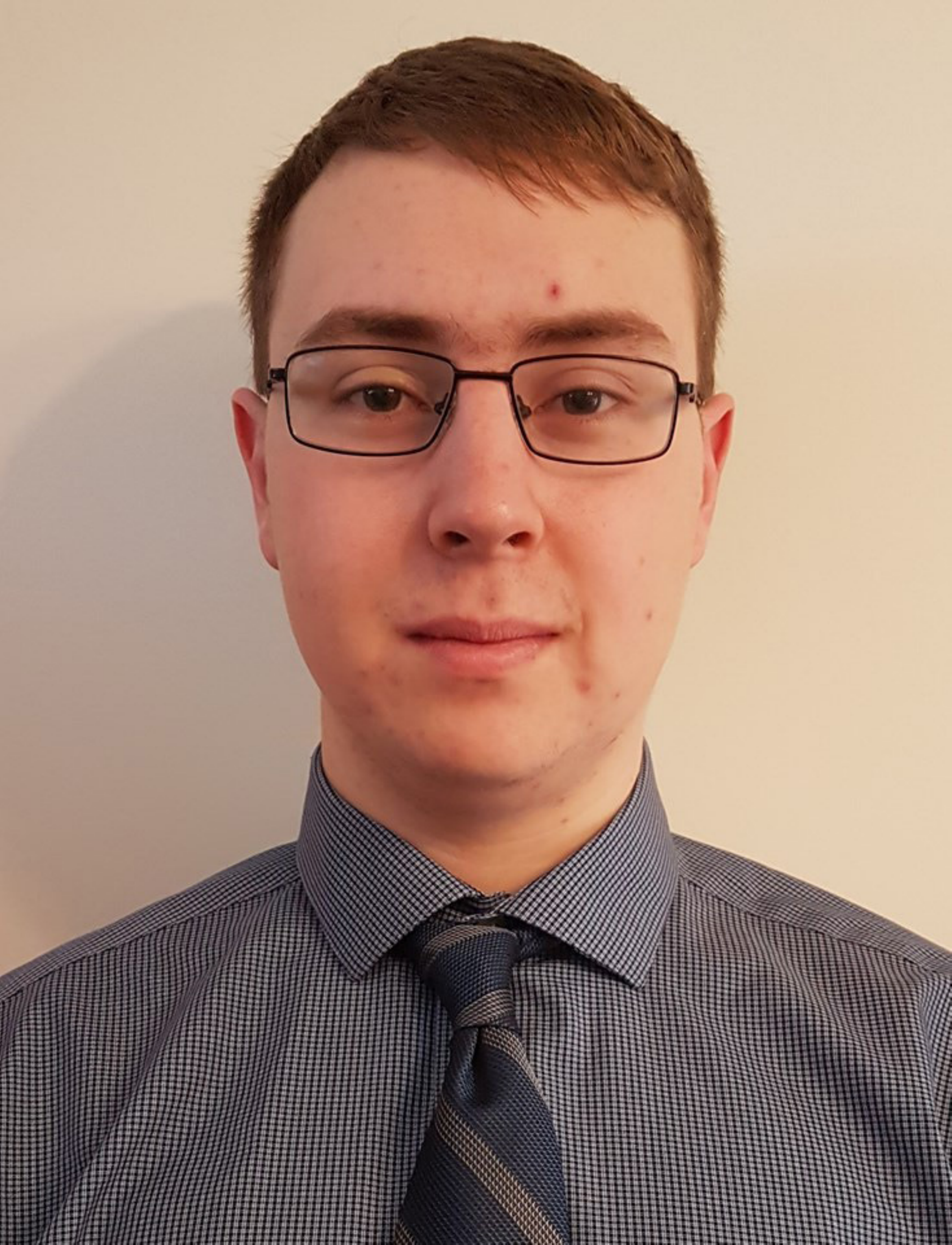}}]{David Chisholm}
	is an undergraduate student who is currently pursuing a bachelors in Electrical Engineering at The University of British Columbia, Okanagan. His research interests include image processing, computer vision, and deep learning.
\end{IEEEbiography}

%
%




\end{document}